\documentclass[review]{elsarticle}

\usepackage{lineno}

\modulolinenumbers[5]

\usepackage{amssymb}
\usepackage{algorithm}
\usepackage{algorithmic}
\usepackage{epsfig}
\usepackage{subfigure}
\usepackage{multirow}
\usepackage{float}
\usepackage{amsmath}
\usepackage{longtable}
\usepackage{booktabs}
\usepackage{subfigure}
\usepackage{fancyhdr}
\usepackage{epstopdf}
\usepackage{ulem}
\usepackage{bm}
\usepackage{amsopn}
\usepackage{amssymb,amsmath,color,times}
\usepackage{amsmath}
\usepackage{amssymb}
\usepackage{amsthm}
\usepackage{bbm}
\usepackage{dsfont}
\usepackage{lineno}
\usepackage{caption}
\usepackage{engord}
\usepackage{lipsum}
\usepackage{float}
\usepackage{graphicx}
\usepackage{color}
\usepackage{txfonts}
\usepackage{pifont}
\biboptions{numbers,sort&compress}
\journal{Computers in Biology and Medicine}
\usepackage[colorlinks,
linkcolor=red,       
anchorcolor=blue,  
citecolor=blue,        
]{hyperref}
\usepackage{cleveref}

\begin{document}
	\begin{frontmatter}
		
		\title{Multi Task Consistency Guided Source-Free Test-Time Domain Adaptation Medical Image Segmentation}
		
		\author[mymainaddress]{Yanyu Ye}
		\author[secondaddress]{Zhenxi Zhang\footnote{{Yanyu Ye and Zhenxi Zhang contribute equally to this paper and co-share the first authorship of this paper.}}}
		\author[mymainaddress]{Wei Wei\corref{mycorrespondingauthor}}
		\cortext[mycorrespondingauthor]{Corresponding author}
		\ead{weiweinwpu@nwpu.edu.cn}
		\author[secondaddress]{Chunna Tian}
		\address[mymainaddress]{School of Computer Science, Northwestern Polytechnical University, Xi'an 710129, China}
		\address[secondaddress]{School of Electronic Engineering, Xidian University, Xi'an 710071, China}

		\begin{abstract}
			Source-free test-time adaptation for medical image segmentation aims to enhance the adaptability of segmentation models to diverse and previously unseen test sets of the target domain, which contributes to the generalizability and robustness of medical image segmentation models without access to the source domain. Ensuring consistency between target edges and paired inputs is crucial for test-time adaptation. To improve the performance of test-time domain adaptation, we propose a multi task consistency guided source-free test-time domain adaptation medical image segmentation method which ensures the consistency of the local boundary predictions and the global prototype representation. Specifically, we introduce a local boundary consistency constraint method that explores the relationship between tissue region segmentation and tissue boundary localization tasks. Additionally, we propose a global feature consistency constraint toto enhance the intra-class compactness. We conduct extensive experiments on the segmentation of benchmark fundus images. Compared to prediction directly by the source domain model, the segmentation Dice score is improved by 6.27\% and 0.96\% in RIM-ONE-r3 and Drishti GS datasets, respectively. Additionally, the results of experiments demonstrate that our proposed method outperforms existing  competitive domain adaptation segmentation algorithms.
		\end{abstract}
		
		
	\end{frontmatter}

	\section{Introduction}
	Source-free test-time domain adaptation method refers to adapting a source model to the test set of the target domain dataset and contributing to the generalizability and robustness of medical image segmentation systems. In typical clinical environments, on one hand, the source domain data is usually inaccessible, and obtaining an abundant training set from the target domain for source model adaptation may be infeasible. Thus, the adaptation can only occur on a few (or even a single) test images of the target domain. On the other hand, there still exist distribution differences between the source data and target data, and even between different test samples of the target domain. These differences can arise from changes in imaging protocols, variations in parameters within the same protocol, inherent hardware differences across machines, and fluctuations in signal-to-noise ratios within the same machine. When using the unlabeled target domain training set for domain adaptation, model performance can deteriorate due to the distribution gap between the training and testing sets. Hence, in the absence of access to the source data and the training set of the target domain, the problem of source-free test time domain adaptation (SFTTA) has become more challenging and garnered attention from many researchers recently.\par
	
	
	Existing SFTTA methods are primarily studied in image classification tasks. Some researchers\cite{sun2020test} \cite{chen2022contrastive} design self-supervised learning methods to achieve source-free test time domain adaptation. These approaches utilize auxiliary tasks\cite{sun2020test}, contrastive learning\cite{chen2022contrastive}, and other techniques to extract knowledge from unlabeled target domain data and fine-tune the source domain pre-trained model. Additionally, some researchers utilize consistency learning methods\cite{ma2022test} to address test-time domain adaptation problems in medical image segmentation tasks. Some researchers\cite{bateson2022test} \cite{karani2021test} introduce prior information to tackle the source-free domain adaptation issue in medical image segmentation tasks. Furthermore, some researchers pay attention to the training strategies during adaptation, including dynamically adjusting the learning rate\cite{yang2022dltta} and selecting neurons to restore the weights from the source pre-training\cite{wang2022continual}. However, these methods neglect the efficient knowledge exploitation from the test samples in the target domain during the testing time, which may lead to sub-optimal performance for test-time adaptation in medical segmentation tasks. In this paper, we design a multi task consistency guided test-time source-free medical image segmentation framework by digging local boundary information and global semantic representation of test samples online.\par
	
	
	In semi-supervised learning, there are two assumptions: The smoothness assumption and the clustering assumption. The smoothness assumption states that similar input data should have similar output. By adding small perturbations to unlabeled data, the predicted results should not change significantly, ensuring output consistency. Numerous studies\cite{yu2019uncertainty}\cite{ouali2020semi}\cite{li2020transformation} have focused on designing unsupervised consistency constraint tasks to extract information from unlabeled samples. Inspired by these works, we propose a local boundary prediction task and a global feature learning task in this paper, which enforces local boundary and global feature consistency, aiming to explore information from the target domain test set to directly adapt the source domain model through training on the target domain test set. \par

	In summary, our proposed method, called multi task consistency guided source-free test-time source-free test time domain adaptation segmentation method (MCDA), provides a new solution for source-free test-time domain adaptation in the context of medical image segmentation. The contributions of this paper are summarized as follows:
	\begin{itemize} 
		\item We design a multi task consistency guided source-free test-time medical image segmentation framework by digging local boundary information and global semantic representation of test samples online.
		\item We propose a boundary consistency constraint method that explores the relationship between tissue region segmentation and tissue boundary localization tasks.
		\item We introduce a target prototype consistency constraint to guide the model's attention towards the target regions in the image, aiming to enhance the model's segmentation performance and adaptability to different contextual information in the target domain at test time.
		\item We evaluate our method on three public fundus image datasets for optic disc and cup segmentation. Without any source data, our method achieves better performance than the state-of-the-art SFTTA method.\par
	\end{itemize}
	

	\label{section:1}
	\section{Related Work}
	We first present related work on source-free domain adaptation and source-free test-time domain adaptation as follows.
	\label{section2}
	\subsection{Source-free domain adaptation}
	\label{section2_1}
	Whether at the feature level or image level, unsupervised domain adaptation methods usually require the utilization of both source domain and target domain data to align the image features or generate cross-domain images. However, in practical applications, the acquisition of source domain data can be challenging due to data privacy concerns. Therefore, the problem of source-free domain adaptation has garnered attention from many researchers.
	
	Some existing source-free domain adaptation works focus on classification tasks. Existing source-free domain adaptation methods for image classification can be mainly categorized into two approaches: Source-like domain generation and pseudo-label self-training. Among them, the source-like domain generation methods generate source-like domain data that resembles the distribution of the source domain data using distribution estimation techniques. Then they transform the source-free domain adaptation problem into an unsupervised domain adaptation problem. For example, Kurmi et al. \cite{kurmi2021domain} proposed the source data free domain adaptation model, where they first use a Generative Adversarial Network (GAN) in conjunction with a pre-trained classifier to learn the underlying data distribution of the source dataset. During training, the source-like data is generated using label values and noise by a class-conditioned GAN. The source-like domain data and target domain data are then jointly used for domain adaptation training. The pseudo-label self-training approach utilizes a source domain pre-trained model to generate pseudo labels for the target domain and adopts an iterative training paradigm for adaptive training. In this process, the pseudo labels are continuously updated, and their quality gradually improves as the iterative training progresses until it reaches stability. Kim et al. \cite{kim2021domain} leverage class prototypes of reliable samples to assign pseudo labels to target samples and reduce the uncertainty of the pseudo-labeling process through distance-based filters. \par
	Compared to image classification, image segmentation tasks are more complex as they require understanding spatial and context information in images. Research and exploration of source-free unsupervised domain adaptation methods in semantic segmentation have been relatively limited, and related works can be broadly categorized into three types: Source-like domain generation, pseudo-label self-training, and regularization constraint introduction.\par
	
	In some typical source-like domain generation approaches, Yang et al. \cite{yang2022source} introduce style loss and content loss to generate class-conditioned source images by constraining the batch normalization (BN) layers and applying Fourier transform. Ye et al. \cite{ye2021source} select high-entropy images as class-conditioned source images and align the distribution of class-conditioned source images with difficult images using adversarial learning.\par
	
	In the pseudo-label self-training methods, obtaining high-quality pseudo labels is crucial. Chen et al. \cite{chen2021source} enhance the pseudo labels by introducing complementary pixel-level and class-level pseudo-label denoising methods. For pixel-level denoising, they use uncertainty estimation to select pseudo labels with higher confidence. For class-level denoising, they calculate the distance between each pixel and the class prototypes of foreground and background, improving the pseudo labels by removing noise. Xu et al. \cite{xu2022denoising} propose the U-DR4 model, which also enhances the quality of pseudo labels through denoising. They use an adaptive class-dependent threshold strategy for rough denoising and then introduce uncertainty-corrected pseudo labels for fine denoising using the estimated joint distribution matrix between observed labels and latent labels. Vs et al. \cite{vs2022target} propose a two-stage method consisting of a specific target adaptation stage and a specific task adaptation stage. In the specific target adaptation stage, the authors generate multiple pseudo labels through image augmentation and further optimize the model by minimizing the information entropy of the pseudo labels. Subsequently, a selective voting method is used to filter out false negatives in the pseudo labels. In the task-specific adaptation stage, strong and weak images are inputted into teacher-student networks for consistent learning.\par
	
	Furthermore, some unsupervised domain adaptation methods have introduced regularization functions. In the works by Vs et al. \cite{vs2022target}, Yang et al. \cite{yang2022source}, and Ye et al. \cite{ye2021source}, consistency regularization is employed during target domain adaptation to align the distribution of target domain data. In the study by Fleuret et al., \cite{fleuret2021uncertainty}, dropout is applied to the decoder parameters to obtain diverse inputs, and then consistency regularization is enforced on multiple predictions to train the network.\par
	
	Additionally, anatomical prior information of the target segmentation can be utilized to guide the unsupervised domain adaptation process. Bateson et al. \cite{bateson2020source} draw inspiration from anatomical knowledge in segmenting spinal images and introduce auxiliary networks to predict target class ratios. During the domain adaptation phase, the KL divergence is used to measure the difference between the target class ratios in the segmented target domain results and the prior knowledge. The network is trained to minimize this difference, enabling the source pre-trained model to adapt to the distribution of the target domain data.\par
	
	However, these methods assume that the training set in the target domain can be used to fine-tune the pre-trained model from the source domain. In the experimental design of this paper, the training set in the target domain is unavailable. Instead, we directly perform test-time adaptation on the pre-trained model from the source domain using the test samples. This approach is more common and challenging in clinical research, as it holds significant implications for the personalized treatment of patients. Many researchers have also shown interest in and conducted research on this issue which we review in the next section.\par
	
	\subsection{Source-free test-time domain adaptation}
	\label{section2_2}
	Existing unsupervised domain adaptation methods for test-time adaptation are mainly applied in image classification tasks. Some researchers have employed self-supervised learning paradigms to achieve test-time adaptation in the absence of source domain labels. Sun et al. \cite{sun2020test} introduced an auxiliary task of predicting image rotation angles to make the model adapt to the test distribution. During test-time training, the auxiliary task shares the feature extraction module with the image classification task, and the model parameters are updated using the loss function imposed by the auxiliary task. Chen et al. \cite{chen2022contrastive} introduced contrastive learning for test-time adaptive image classification. The model utilizes the MoCo \cite{he2020momentum} contrastive learning framework, where augmented images serve as positives and different images serve as negatives, optimizing the model by minimizing the distance between positive features and maximizing the distance between negative features. Liu et al. \cite{liu2022vmfnet} modeled the composition components of human anatomy as learnable von Mises Fisher kernels and utilized kernels with robustness to different domain images to extract features for image reconstruction and classification.\par
	
	Test-time domain adaptation in medical image segmentation has also received attention from researchers. Wang et al. \cite{wang2020tent} addressed the test-time adaptation problem in the absence of labeled target data by minimizing the entropy of test set predictions. Bateson \cite{bateson2022test} proposed a shape-guided entropy minimization loss for test-time adaptation. They computed shape statistics, such as centroids and centroid distances based on predicted labels, and used KL divergence between these features and the average centroids and centroid moments of the entire test set to guide the model's adaptation training. Karani et al. \cite{karani2021test} employed a separately trained denoising autoencoder module that modeled an implicit prior for anatomical segmentation labels. During testing, the image normalization module was adaptively trained under the guidance of the implicit prior, and the normalized image segmentation module was used to generate predicted segmentation labels. Yang et al. \cite{yang2022dltta} argued that previous test-time adaptation methods have a common limitation. They use a fixed learning rate during adaptation training. Test data exhibits varying degrees of distribution shift in practical applications, rendering the training using the fixed learning rate suboptimal. To address this issue, they propose a dynamic learning rate adjustment method for test-time adaptation, which dynamically adjusts the weight update magnitude for each test image to alleviate differences in distribution shift. To prevent catastrophic forgetting during the adaptation process, Wang et al. \cite{wang2022continual} proposed randomly restoring a small portion of neurons to the weights of the source pre-training at each iteration, aiming to preserve the source knowledge in the long term. However, these efforts have overlooked the potential of utilizing consistency constraints to extract prior information for adaptation. In this paper, we propose a multi task consistency guided source-free test-time medical image segmentation framework by digging local boundary information and global semantic representation of test samples online.\par

	\section{Methodology}
	\label{section3}
	First, we introduce an overview in subsection \ref{subsection3.1}. Then, we present the pretraining model in subsection \ref{subsection3.2}. Subsequently, for source-free test-time domain adaptation, we introduce local boundary consistency constraint and global feature consistency constraint in subsection \ref{subsection3.3} and \ref{subsection3.4}, respectively. Finally, we discuss the loss function in subsection \ref{subsection3.5}.
	
	\subsection{Overview}
	\label{subsection3.1}
\begin{figure}[h]
		\centering\includegraphics[scale=0.36]{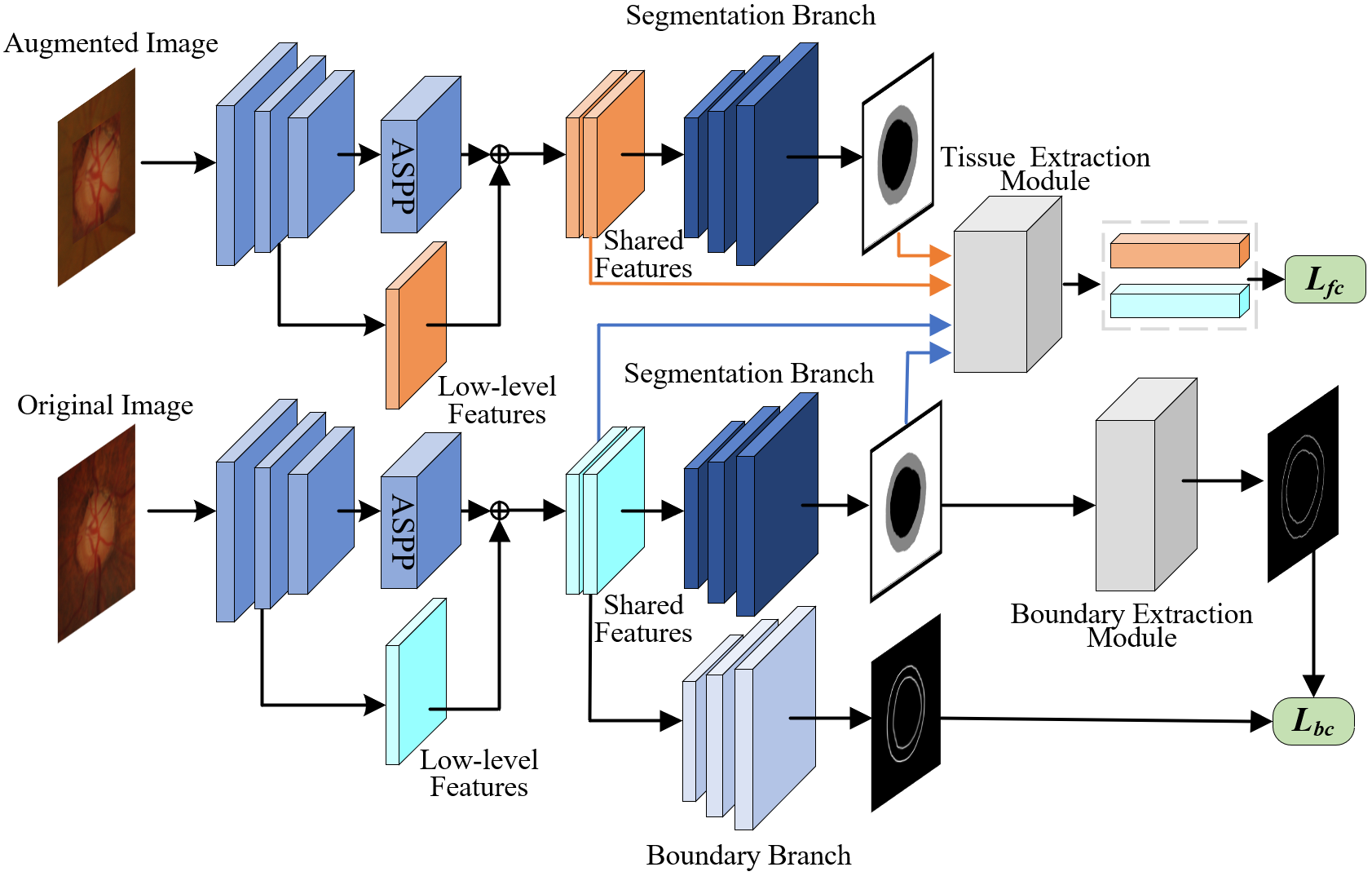}\caption{The overall framework of  the proposed MCDA method}\label{TCDA_model}
	\end{figure}

	The source domain training dataset, denoted as ${D_S} = { (x_i^S,y_i^S) \in ({X_S},{Y_S})} _{i = 1}^{{N_1}}$, consists of pairs of images and corresponding labels in the source domain. Here, $x_i^S$ represents the $i$-th original image in the source domain, and $y_i^S \in {[0,1]^{H \times W \times C}}$ denotes the label values for the optic disc and cup in the source domain. The values of $H$ and $W$ represent the length and width of the images and labels, while $C$ represents the number of classes. In the label values, if a pixel has a value of 1 for a particular class, it indicates that the pixel belongs to the foreground of that class. Conversely, if the label value is 0, it indicates that the pixel belongs to the background. In the context of optic cup and optic disc segmentation datasets, $C$ is set to 2, representing the classes of optic cup and optic disc. The unlabeled target domain test set, denoted as ${D_T} = { (x_i^T)} _{i = 1}^{{N_2}}$, contains unannotated images from the target domain. The source domain pre-trained model is denoted as ${\phi _S}(\cdot)$, and the adaptively trained target domain model is denoted as ${\phi _T}(\cdot)$. Fig.\ref{TCDA_model} illustrates the framework of the proposed MCDA model. 

	\subsection{Pretrained model in source domain}
	\label{subsection3.2}
	\par
\begin{figure}[h]
		\centering\includegraphics[scale=0.4]{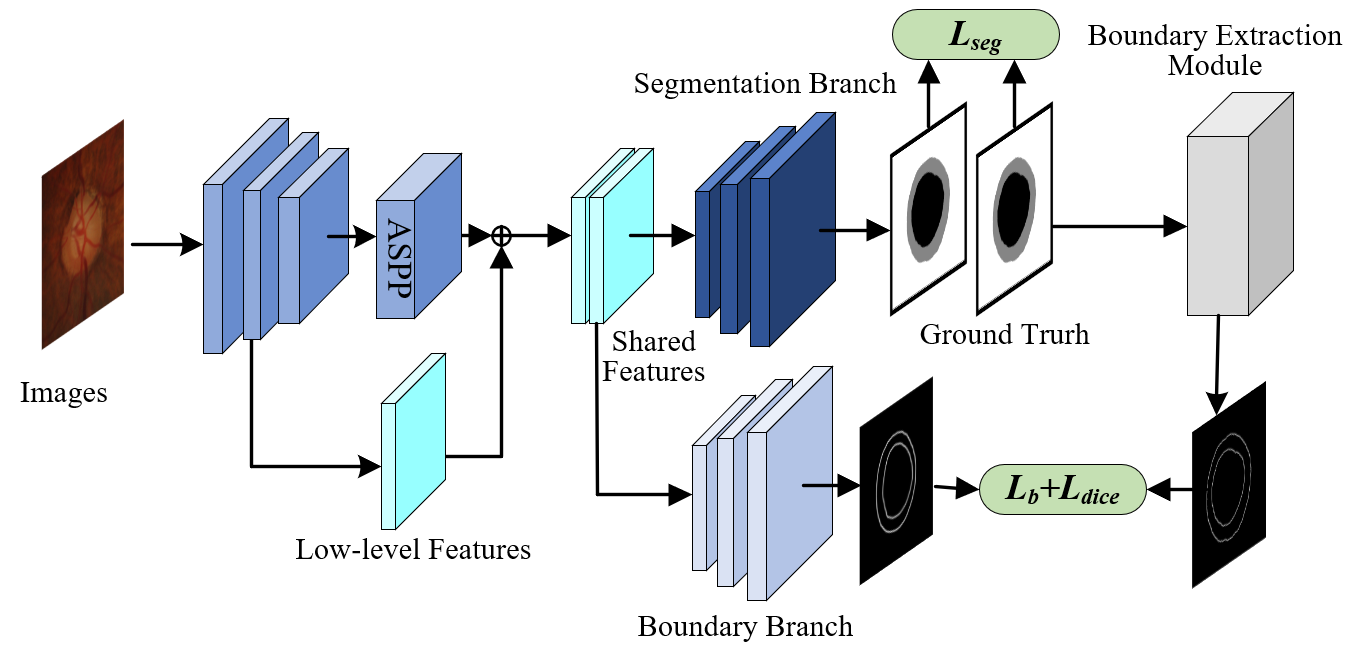}\caption{The framework of pre-trained model on source domain. We introduce an additional boundary branch based on the shared features. In addition, we design a boundary prediction consistency constraint by $L_b$ and $L_{dice}$.}\label{source_model}
	\end{figure}

	We employ DeepLab v3+ as the segmentation network. We introduce a boundary prediction branch into the original DeepLab v3+ network, forming a multi-task network. The overall architecture is shown in Fig. \ref{source_model}.
	
	For each image $x_i^S$ in the source domain dataset, we feed it to the multi-task DeepLab v3+ network and get segmentation results and boundary prediction results. This process is defined as follows:
	\begin{equation}
		\hat y_i^S,\hat b_i^S = {\phi _S}(x_i^S)\label{eq31}
	\end{equation}
	\noindent
	$\hat y_i^S$ represents the segmentation results from the segmentation prediction branch, while $\hat b_i^S$ represents the predicted tissue boundaries from the boundary prediction branch.
	
	To obtain the boundary labels of the source domain images, we utilize the boundary extraction algorithm based on the Sobel operator. The Sobel operator template, as shown in Eq.(\ref{eq32}), consists of a 3$\times$3 matrix where ${d_x}$ represents the horizontal direction and ${d_y}$ represents the vertical direction.
	\begin{equation}
		{d_x} = \left[ {\begin{array}{*{20}{c}}
				{ - 1}&0&1\\
				{ - 2}&0&2\\
				{ - 1}&0&1
		\end{array}} \right],{d_y} = \left[ {\begin{array}{*{20}{c}}
				{ - 1}&{ - 2}&{ - 1}\\
				0&0&0\\
				1&2&1
		\end{array}} \right]\label{eq32}
	\end{equation}
	
	Next, we obtain the boundary labels for the source domain images based on Eq.(\ref{eq33}), where $y_i^S$ represents the label of the source domain image $x_i^S$.
	\begin{equation}
		b_i^S = sobel(y_i^S)\label{eq33}
	\end{equation}
	
	For the segmentation branch, we utilize cross-entropy loss $L_s$ to optimize the training process, which is defined as: 
	\begin{equation}
		{L_{seg}} =  - \sum\limits_{i \in {N_1}} {y_i^S\log (\hat y_i^S) + (1 - y_i^S)\log (1 - \hat y_i^S)} 
	\end{equation}
	\noindent
	$y_i^S$ represents the ground truth label for the $i$-th image in the source domain dataset, and $\hat{y}_i^S$ represents the predicted result of the segmentation branch for the $i$-th image.
	
	For the boundary prediction branch, we first utilize the cross-entropy loss function to train the network to approximate the target boundary prediction. Then, we refine the network's boundary prediction using the Dice loss function.
	\begin{equation}
		{L_{b}} =  - \sum\limits_{i \in {N_1}} {b_i^S\log (\hat b_i^S) + (1 - b_i^S)\log (1 - \hat b_i^S)} 
	\end{equation}
	\begin{equation}
		{L_{dice}} = 1 - \frac{{2|b_i^S \cap \hat b_i^S|}}{{|b_i^S| + |\hat b_i^S|}}
	\end{equation}
	
	\begin{equation}
		{L_{boundary}} = \left\{ {\begin{array}{*{20}{c}}
				{{L_{b}},epoch \le 1000}\\
				{{L_{dice}},1000 < epoch \le 1200}
		\end{array}} \right.
	\end{equation}
	
	\noindent
	$b_i^S$ represents the ground truth boundary labels for the $i$-th image in the source domain dataset, and $\hat{b}_i^S$ represents the predicted result of the boundary prediction branch for the $i$-th image. The loss function for training the source domain model is defined as $L_S$.
	\begin{equation}
		{L_S} = {L_{seg}} + {L_{boundary}}
	\end{equation}
	
	\subsection{Local boundary consistency constraint}
	\label{subsection3.3}
	Region-based segmentation \cite{guo2022joint} \cite{li2021medical}methods highlight the global homogeneity of pixel semantic information and object-level contextual information. While boundary-based segmentation methods\cite{kervadec2019boundary} \cite{chen2021deep} focus on local boundary features and spatial variations on both sides of the boundary contour. When segmenting one image, region segmentation methods, and boundary segmentation methods capture different information from the image. Notably, the boundaries extracted from segmentations should be consistent with the results of boundary predictions. Therefore, we introduce a local boundary consistency constraint to optimize the network, ensuring that the network forms tissue boundary consistency and adapts the source domain model to the distribution of the test set of the target domain.
	
	Initially, the target domain test set data is fed into the source domain pre-trained model, yielding predictions for region segmentation and boundary detection.
	\begin{equation}
		\hat y_i^T,\hat b_i^T = {\phi _T}(x_i^T)
	\end{equation}
	\noindent
	$\hat y_i^T \in \mathbb{R}^{{H} \times {W} \times {C}}$ represents the segmentation prediction, and $\hat b_i^T \in \mathbb{R}^{{H} \times {W} \times {C}}$ represents the boundary prediction. ${H}$, ${W}$, and ${C}$ denote the dimensions of the prediction (height, width, and number of classes, respectively). ${\phi _T}(\cdot)$ represents the target domain model, which is initialized as ${\phi _S}(\cdot)$ at the beginning of training.
	
	Next, the boundaries are extracted from the segmentation predictions using the same method as described in the previous subsection \ref{subsection3.2}, as shown in Eq.(\ref{eq315}).
	
	\begin{equation}
		\tilde b_i^T = sobel(\hat y_i^T)\label{eq315}
	\end{equation}
	\noindent
	$\tilde b_i^T \in \mathbb{R}^{{H} \times {W} \times {C}}$ represents the boundaries extracted from the segmentation predictions.

	To ensure consistency between the boundaries obtained from segmentation predictions and the results of boundary predictions for the same image, we introduce a consistency loss computed using the L2 norm, which enables the source pre-trained model to adapt to the distribution of the test data from the target domain. The formula of the consistency loss is defined as follows:
	\begin{equation}
		{L_{bc}} = ||\hat b_i^T-\tilde b_i^T
		|{|_2}\label{eq319}
	\end{equation}
	\noindent
	where $||\cdot|{|_2}$ denotes the L2 norm. \par
	
	\subsection{Global feature consistency constraint}
	\label{subsection3.4}
	Feature capability plays a crucial role in deep learning-based segmentation. In the task of unsupervised test-time domain adaptation, directly training the model on the target domain test set can lead to issues such as overfitting to a few images and excessive reliance on contextual information for segmentation, especially when the dataset is small. To address these challenges and improve the performance of the model in adapting to the target domain test images, we propose a global feature consistency constraint.
	

	Firstly, the test images from the target domain are fed into the source model, generating the pseudo labels of the target domain test set.
	\begin{equation}
		p_i^T = {\phi _S}(x_i^T)
	\end{equation}
	\noindent
	Let $p_i^T \in {\mathbb{R}^{{H} \times {W} \times {C}}}$ represent the pseudo labels for the $i$-th image in the target domain test set.
	
	To crop the tissue region from the image, we initiate a top-down scan to acquire rectangles smallest encompassing the target. The coordinates of this rectangle's top-left and bottom-right corners are denoted as ($h_1$, $w_1$) and ($h_2$, $w_2$), respectively. This process yields an image $x_{{\rm{target,i}}}^T$ which contains only the tissue region. $x_{{tissue,i}}^T \in {\mathbb{R}^{{H_1} \times {W_1} \times {C_1}}}$, where ${C_1}$ is the number of image channels, and ${H_1} = h_2 - h_1$, ${W_1} = w_2- w_1 $.

	Similarly, we employ the same method to extract another image $x_j^T$ that solely contains the tissue region. The coordinates of the top-left and bottom-right corners of this rectangle are denoted as ($h_3$, $w_3$) and ($h_4$, $w_4$). The dimensions of the corresponding rectangle are denoted as ${H_2}$ and ${W_2}$.To replace the background in $x_{{target,i}}^T$ while ensuring that the original target is not included in the background, we resize $x_{{target,i}}^T$ to match the dimensions of ${H_2}$ and ${W_2}$.
	\begin{equation}
		x_{{target,i}}^T = resize(x_{{target,i}}^T)
	\end{equation}
	
\begin{figure}[h]
		\centering\includegraphics[scale=0.25]{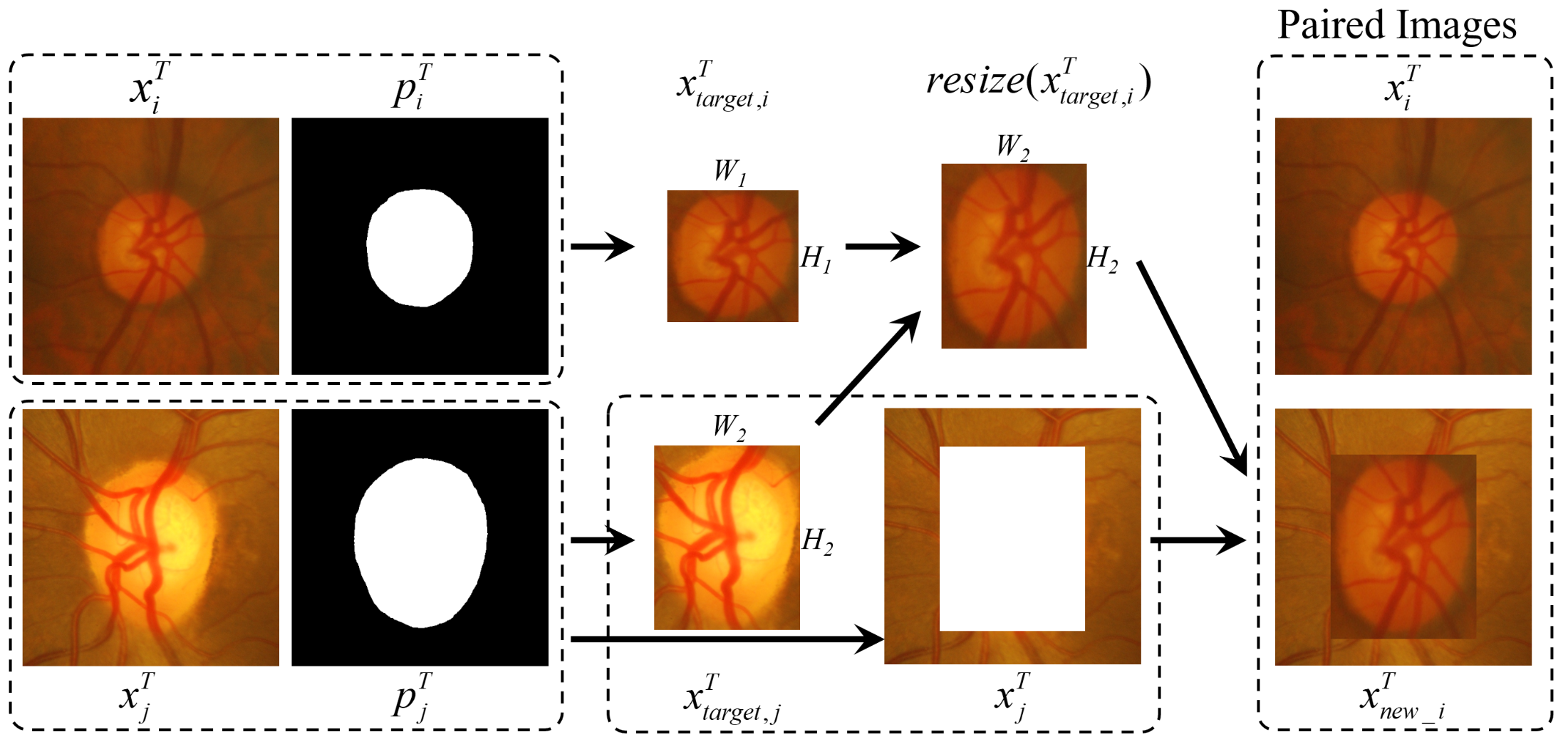}\caption{The demonstration of target domain image augmentation}\label{mix}
	\end{figure}

	Subsequently, we obtain the replaced background image $x_{{new\_}i}^T$ as shown in Fig.\ref{mix}. By obtaining  $x_{{new\_}i}^T$, we achieve a modified version of the target domain image that focuses on the tissue region while ensuring a different background context appearance of another test image.
	Fig.\ref{mix} demonstrates the entire process of background replacement in the images. During the training process, within the same batch of $x_i^T$, a random image $x_j^T$ is selected as the background image. 
	
	When the network focuses on segmenting the target while avoiding excessive reliance on contextual information in the image, it is expected that the same target with different contextual information should generate similar target features. To utilize this cue for better test-time adaptation, the original image $x_i^T$ and the image with the replaced background $x_{{new\_}i}^T$ are simultaneously fed into the model. 
	\begin{equation}
		\hat y_i^T,f_i^T = {\phi _T}(x_i^T)\label{eq316}
	\end{equation}
	\begin{equation}
		\hat y_{new\_i}^T,f_{new\_i}^T = {\phi _T}(x_{new\_i}^T)\label{eq313}
	\end{equation}
	
	By applying Eq.(\ref{eq316}) and Eq.(\ref{eq313}), their predicted segmentation map $\hat y_i^T$ and $\hat y_{new\_i}^T$, and intermidiate features $f_i^T$and $f_{new\_i}^T $ can be obtained. Let $\hat y_i^T$, $\hat y_{new\_i}^T \in \mathbb{R}^{H \times W \times C}$ represent the predicted segmentation maps of the $i$-th image and its corresponding background-replaced image, respectively. On the other hand, $f_i^T$ and $f_{new\_i}^T \in \mathbb{R}^{H \times W \times C_2}$ represent the pixel-level feature of the $i$-th image and its corresponding background-replaced image, respectively. These features contain important semantic information about the tissue objects in the images. $C_2$ is the channel of the feature.
	
	We compute the tissue feature prototypes for the paired images in Eq.(\ref{eq37}) and (\ref{eq38}),
	\begin{equation}
		{f_{obj,new\_i}} = \sum\nolimits_j {\mathbbm{1}[\hat y_{new\_i}^T(j) =  = 1] \times f_{new\_i}^T(j)} /\sum\nolimits_j {\mathbbm{1}[} \hat y_{new\_i}^T(j) = = 1]\label{eq37}
	\end{equation}
	\begin{equation}
		{f_{obj,i}} = \sum\nolimits_j {\mathbbm{1}[\hat y_{\rm{i}}^T(j) =  = 1] \times f_{i,j}^t} /\sum\nolimits_j {\mathbbm{1}[} \hat y_i^T(j) =  = 1]\label{eq38}
	\end{equation}
	
	\noindent
	${f_{obj,i}}$ represents the foreground tissue prototype for the $i$-th image, ${f_{obj,new_i}}$ represents the foreground tissue prototype for the $i$-th image with the replaced background. By using the cup mask or disc mask for prototype calculation, we can obtain the disc prototype vectors ${f_{disc,i}}$ and ${f_{disc,new_i}}$, as well as the cup prototype vectors ${f_{cup,i}}$ and ${f_{cup,new_i}}$.
	
	To encourage similarity among features of the same target, we introduce the feature consistency loss for both the optic cup and disc. The cosine distance loss function is used to ensure the consistency between different global features of the tissue, thus making the source model adaptive to the test images in the target domain. 
	\begin{equation}
		{L_{cup}} = 1 - f_{cup,i}^T \cdot f_{cup,new\_i}^T/|f_{cup,i}^T| \times |f_{cup,new\_i}^T|\label{eq39}
	\end{equation}
	\begin{equation}
		{L_{disc}} = 1 - f_{disc,i}^T \cdot f_{disc,new\_i}^T/|f_{disc,i}^T| \times |f_{disc,new\_i}^T|\label{eq310}
	\end{equation}
	Finally, we define the loss of global feature consistency constraint ${L_{fc}}$  in Eq.(\ref{eq311}).
	\begin{equation}
		{L_{fc}} = {L_{cup}} + {L_{disc}}\label{eq311}
	\end{equation}
	
	\subsection{Overall loss function}
	\label{subsection3.5}
	During the test-time domain adaptation for retinal cup-disc segmentation in the absence of source domain data, we optimize the model using the target boundary consistency loss and target feature consistency loss to adapt the model to the data distribution of the target domain test set. However, the distribution discrepancy can lead to the problem of pattern collapse during the consistency learning process. To address this issue and prevent performance degradation of the source pre-trained model. We introduce a segmentation loss. For calculating the segmentation loss, we utilize the pseudo labels generated by the source model as the target labels for the target domain test set and compute the binary cross-entropy loss ${L_{Tseg}}$.
	\begin{equation}
		{L_{Tseg}} =  - \sum\limits_{x_i^T \in {X_T}} {p_i^T\log (\hat y_i^T)} 
	\end{equation}
	\noindent
	$p_i^T $ represents the pseudo-label obtained directly from the source domain model.
	
	In our approach, the loss function consists of three components: Target boundary consistency loss ${L_{bc}}$, target feature consistency loss ${L_{fc}}$, and segmentation loss ${L_{Tseg}}$. The overall loss function is formulated as Eq.(\ref{eq312}).
	\begin{equation}
		{L_{total}} = {L_{Tseg}} + \alpha{\sum\limits_{x_i^T \in {X_T}} {{L_{bc}}}} + \beta {\sum\limits_{x_i^T \in {X_T}} {{L_{fc}}}} \label{eq312}
	\end{equation}
	\noindent
	$\alpha$ and $\beta$ are hyperparameters in our model.

	\section{Experiment Settings}
	\label{section4}
	\subsection{Dataset}
	We use three datasets from different sites to validate the effectiveness of the proposed LGDA. REFUGE\cite{orlando2020refuge} dataset is regarded as the source domain and RIM-ONE-r3\cite{fumero2011rim} and Drishti-GS\cite{sivaswamy2015comprehensive} datasets are treated as target domains. The source domain includes 400 annotated training images, and the two target domain includes 60 and 51 test images, respectively. The data preprocessing of this paper follows the setting in the literature\cite{chen2021source}. The fundus image is cropped into a region of interest (ROI) centered on the optic disc as the network input with the size of 512$\times$512. Additionally, we use the common data augmentation strategies including random rotation, flipping, elastic transformation, contrast adjustment, adding Gaussian noise, and random erasing.
	
	\subsection{Evaluation metric}
	For evaluation, we employ two commonly used metrics, including the Dice coefficient for overlap measurement and average surface distance (ASD) for boundary consistency evaluation.  \par
	The Dice coefficient is used to describe the proportion of the overlapping region between the predicted result and the ground truth annotation in the overall image area. Higher Dice indicates better performance.
	\begin{equation}
		Dice = \frac{2(y_i^T \cap \hat y_i^T)}{(|y_i^T| + |\hat y_i^T|)}\label{eq3-1}
	\end{equation}
	\noindent 
	Eq.(\ref{eq3-1}) represents the calculation equation for the Dice coefficient, where $y_i^T$ stands for the ground truth annotation of the segmentation result, and $\hat y_i^T$ represents the output generated by the network prediction.
	
	The Average Surface Distance (ASD) refers to the average distance from all points within an object to its surface. A smaller distance indicates a closer alignment between the predicted result and the bounding surface of the ground truth annotation. The ASD is defined as Eq.(\ref{eq3-2}). Let $j$ denote the index of the pixel. 
	\begin{equation}
		ASD = \frac{1}{{|S(\hat y_i^T)| + |S(y_i^T)|}}(\sum\nolimits_{S(\hat y_{i,j}^T) \in S(\hat y_i^T)} d (S(\hat y_{i,j}^T) ,S(y_i^T)) + \sum\nolimits_{S(y_{i,j}^T) \in S(y_i^T)} d (S(y_{i,j}^T),S(\hat y_i^T)))\label{eq3-2}
	\end{equation}
	\noindent
	where $d (S(\hat y_{i,j}^T) ,S(y_i^T))$ refers to the Euclidean distance from boundary pixels in the prediction $S(\hat y_{i,j}^T)$ to the nearest pixel in the boundary of the ground truth $S(y_i^T)$, which is formulated as:
	\begin{equation}
		d (S(\hat y_{i,j}^T) ,S(y_i^T)) = \mathop {\min }\limits_{{S(y_{i,j}^T)} \in S(y_i^T)} ||{S(\hat y_{i,j}^T)} - {S(y_{i,j}^T)}||
	\end{equation}

	\subsection{Implementation details}
	We employ the DeepLab v3+ network as the backbone segmentation network. All methods are implemented in PyTorch and trained on one NVIDIA TITAN RTX GPU. The batch size is set to 8. We use the Adam optimizer in our experiments. We set the fixed learning rate to 0.001. All experiments follow the same training settings.

	\section{Experimental results and analysis}
	\label{section5}
	\subsection{Comparison with State-of-the-Arts}
	We compare our method with recent state-of-the-art domain adaptation methods, including BEAL\cite{wang2019boundary}, AdvEnt\cite{vu2019advent}, FSM\cite{yang2022source}, DPL\cite{chen2021source}, DAE\cite{karani2021test}, Tent\cite{wang2020tent} and CoTTA\cite{wang2022continual}. Among them, the BEAL\cite{wang2019boundary} and AdvEnt\cite{vu2019advent} are unsupervised domain adaptation methods. Boundary information is also used for adaptation in the BEAL\cite{wang2019boundary}. The FSM\cite{yang2022source} and DPL\cite{chen2021source} are source-free domain adaptive models trained with the target domain training set. DAE\cite{karani2021test}, Tent\cite{wang2020tent}, and  CoTTA\cite{wang2022continual} are source-free test-time adaptation methods. In Table \ref{RIM-ONE-r3数据集TCDA} and Table \ref{Drishti-GS数据集TCDA} we label the unsupervised domain adaptation methods with "U", the source-free domain adaptation methods with "F", and the source-free domain test adaptive methods with "T". Additionally, "w/o adaptation" represents the results obtained by directly testing the source domain model on the target domain test set, while "Upper bound" denotes the results achieved by training a model directly on the target domain training set and subsequently testing it.\par
	
	Table \ref{RIM-ONE-r3数据集TCDA} and Table \ref{Drishti-GS数据集TCDA} report the comparisons with other popular methods on the RIM-ONE-r3 dataset and Drishti GS dataset, respectively. 
	
	\begin{table}[H]
		\renewcommand{\arraystretch}{1.2}
		\centering
		\caption{Comparison of experimental results of MCDA model in RIM-ONE-r3 dataset}
		\label{RIM-ONE-r3数据集TCDA}
		\resizebox{\linewidth}{!}{{\begin{tabular}{ccccccc}
					\hline
					\multirow{2}{*}{Method} & \multicolumn{2}{c}{Optic disc segmentation} & \multicolumn{2}{c}{Optic cup segmentation} & \multicolumn{2}{c}{Avg}\\
					& Dice $[\% ]$ & ASD(piexl) & Dice $[\% ]$ & ASD(piexl)& Dice $[\% ]$ & ASD(piexl)\\
					\hline
					w \textbackslash o adaptation  & 85.96$ \pm $5.32  & 13.48$ \pm $5.57  & 74.95$ \pm $19.04 & 10.58$ \pm $5.87 & 80.46 & 12.02 \\
					Upper bound   & 94.74$ \pm $2.16  & 4.43$ \pm $1.70  & 80.58$ \pm $20.92 & 7.06$ \pm $6.83  & 87.66 & 5.75  \\
					\hline
					BEAL\cite{wang2019boundary}(U)   & 88.70$ \pm $3.53  & 16.63$ \pm $5.58  & 79.00$ \pm $2.29 & 14.49$ \pm $6.78  & 83.85 & 15.56 \\
					AdvEnt\cite{vu2019advent}(U) & 89.73$ \pm $3.66  & 9.84$ \pm $3.86  & 77.99$ \pm $21.08 & 7.57$ \pm $4.24   & 83.86 & 8.71  \\
					DPL\cite{chen2021source}(F)    & 89.47$ \pm $4.56  & 6.92$ \pm $8.24   & 81.93$ \pm $14.96 & 9.56$ \pm $3.57   & 85.70 & 8.24  \\
					FSM\cite{yang2022source}(F) & 84.42$ \pm $4.19 & 16.53$ \pm $9.44 & 80.14$ \pm $13.28 & 8.33$ \pm $4.70 & 82.28 & 12.43 \\
					\hline
					DAE\cite{karani2021test}(T)    & 89.09$ \pm $3.32  & 11.63$ \pm $6.84  & 79.01$ \pm $12.82 & 10.31$ \pm $8.45  & 84.05 & 10.97 \\
					Tent\cite{wang2020tent}(T)   & 82.93$ \pm $8.95  & 20.76$ \pm $14.32 & 77.03$ \pm $19.04 & 11.21$ \pm $10.61 & 80.12 & 15.99 \\
					CoTTA\cite{wang2022continual}(T)   & 88.57$ \pm $3.85  & 11.17$ \pm $4.86 & 78.16$ \pm $21.10 & 9.59$ \pm $11.96 & 83.36 & 10.38 \\
					MCDA(T)   & \textbf{90.58$ \pm $10.99} & \textbf{10.99$ \pm $9.72}  & \textbf{82.87$ \pm $13.15} & \textbf{7.32$ \pm $5.27}   & \textbf{86.73} & \textbf{9.16} \\ \hline
			\end{tabular}}
		}
		
	\end{table}

	According to the results presented in Table \ref{RIM-ONE-r3数据集TCDA}, our model achieves an average Dice coefficient of 86.73\% and an average ASD coefficient of 9.16 on the RIM-ONE-r3 dataset. In comparison to the second-best DAE model, our model demonstrates a 2.65\% improvement in the Dice coefficient and a 1.81 decrease in the ASD coefficient. The highest Dice score and the lowest ASD score indicate that the proposed model, by incorporating local boundary consistency constraints and global feature consistency constraints, can effectively adapt the source domain pre-trained model to the data distribution of the test set of the target domain without the requirement of training data in the target domain, thereby enhancing the test-time domain adaptation performance.
	
	For the optic disc segmentation task, the MCDA model achieves a Dice score of 90.58\% and an ASD score of 10.99, which outperforms the second-best DAE by 1.49\% on Dice, demonstrating the effectiveness of MCDA on the test-time DA segmentation task. At the same time, there is a decrease of 0.64 in the ASD coefficient compared to the DAE model. The local boundary consistency constraint and global feature consistency constraint that we have introduced offer significant advantages in enhancing boundary segmentation capabilities (inter-class separability) and global feature representation (intra-class feature compactness) during test-time domain adaptation. Consequently, these constraints contribute to a notable improvement in segmentation performance. For the optic cup segmentation task, the proposed model achieves a Dice score of 82.87\% and an ASD score of 7.32. Compared to the second-best DAE model, the Dice score is improved by 3.86\% and the ASD score is decreased by 2.99. This demonstrates the effectiveness of local boundary consistency learning and global feature consistency learning for optic cup segmentation.
	
	In addition, our method outperforms existing popular SFDA methods which need the target domain training set for adaptive training. Our proposed MCDA model is 1.11\% and 0.94\% higher than the best SFDA performance achieved by DPL in optic cup and optic disc segmentation, respectively. This finding illustrates that our approach, which involves multi task consistency guided source-free medical image segmentation method, proves to be highly effective in boosting test-time domain adaptation segmentation performance even in the absence of target domain training data.
	
	In the Drishti-GS dataset, our method achieves the best average Dice scores and average ASD coefficients, as presented in Table \ref{Drishti-GS数据集TCDA}. Specifically, the MCDA model achieves an average Dice score of 91.27\% and an average ASD coefficient of 6.64 for the tasks of fundus
	image segmentation. In the optic disc segmentation task, our model achieves a Dice score of 96.02\% and an ASD coefficient of 4.56. Compared to the CoTTA model, our model achieves the same Dice coefficient, while reducing the ASD coefficient by 0.04. Moreover, for the optic cup segmentation task, our model achieves a Dice coefficient of 86.51\% and an ASD coefficient of 7.31, superior to existing state-of-the-art test-time adaptation methods. The MCDA model outperforms existing source-free test-time adaptation methods in terms of Dice score and achieves the best ASD coefficient. The superior performance of the MCDA model in the optic disc and optic cup segmentation tasks demonstrates our method's efficacy in enhancing the segmentation performance of the source domain model on the target domain test set while maintaining stability. Furthermore, the results presented in Table \ref{Drishti-GS数据集TCDA} indicate that the MCDA model's segmentation results outperform other popular SFDA methods and UDA methods, suggesting that our approach effectively overcomes the adverse conditions of missing source domain data and target domain training data.
	
	\begin{table}[H]
		\renewcommand{\arraystretch}{1.2}
		\centering
		\caption{Comparison of experimental results of MCDA model in Drishti GS dataset}
		\label{Drishti-GS数据集TCDA}
		\resizebox{\linewidth}{!}{{	\begin{tabular}{ccccccc}
					\hline
					\multirow{2}{*}{Method} & \multicolumn{2}{c}{Optic disc segmentation} & \multicolumn{2}{c}{Optic cup segmentation} & \multicolumn{2}{c}{Avg}\\
					& Dice $[\% ]$ & ASD(piexl) & Dice $[\% ]$ & ASD(piexl)& Dice $[\% ]$ & ASD(piexl)\\
					\hline
					w \textbackslash o adaptation  & 96.66$ \pm $1.12 & 3.78$ \pm $1.34 & 81.55$ \pm $11.94 & 11.94$ \pm $7.86 & 89.10 & 7.86 \\
					Upper bound   & 96.65$ \pm $1.60  & 3.60$ \pm $1.50  & 89.09$ \pm $11.23 & 6.78$ \pm $3.68  & 92.87 & 5.19  \\
					\hline
					BEAL\cite{wang2019boundary}(U)   & 95.54$ \pm $2.09 & 7.78$ \pm $3.37 & 85.95$ \pm $11.44 & 14.51$ \pm $8.15 & 90.75 & 11.14 \\
					AdvEnt\cite{vu2019advent}(U) & 96.16$ \pm $1.65 & 4.36$ \pm $1.83 & 82.75$ \pm $11.08 & 11.36$ \pm $7.22 & 89.46 & 7.86  \\
					DPL\cite{chen2021source}(F)    & 96.53$ \pm $1.29 & 3.92$ \pm $1.43 & 83.15$ \pm $11.78 & 11.42$ \pm $6.56 & 89.84 & 7.67  \\
					FSM\cite{yang2022source}(F) & 95.85$ \pm $2.36 & 4.67$ \pm $2.47 & 82.24$ \pm $13.30 & 12.03$ \pm $6.56 & 89.04 & 8.35 \\
					\hline
					DAE\cite{karani2021test}(T)    & 94.04$ \pm $2.85 & 8.79$ \pm $7.45 & 83.11$ \pm $+11.89 & 11.56$ \pm $6.32 & 88.58 & 10.18 \\
					Tent\cite{wang2020tent}(T)   & 94.73$ \pm $2.32 & 7.53$ \pm $7.79 & 85.76$ \pm $11.12 & 9.88$ \pm $6.32  & 89.86 & 8.71  \\
					CoTTA\cite{wang2020tent}(T)   & \textbf{96.02$ \pm $1.44} & 4.60 $ \pm $1.79 & 83.73$ \pm $11.36 & 10.77$ \pm $6.15  & 89.88 & 7.68  \\
					MCDA(T)   & 96.02$ \pm $1.75 & \textbf{4.56$ \pm $1.96} & \textbf{86.51$ \pm $12.13} & \textbf{8.71$ \pm $5.09}  & \textbf{91.27} & \textbf{6.64}  \\ \hline
			\end{tabular}}
		}
	\end{table}
	
	\subsection{Visualization}
\begin{figure}[h]
		\centering\includegraphics[scale=0.8]{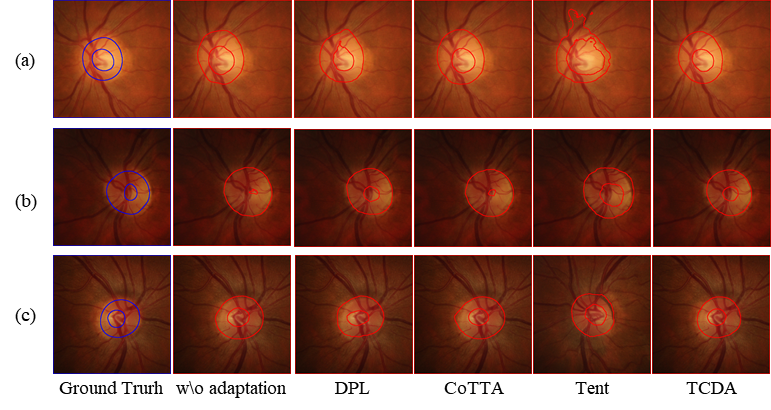}\caption{Visual segmentation results for samples on the RIM-ONE-r3 dataset}\label{visR}
	\end{figure}

\begin{figure}[h]
		\centering\includegraphics[scale=0.8]{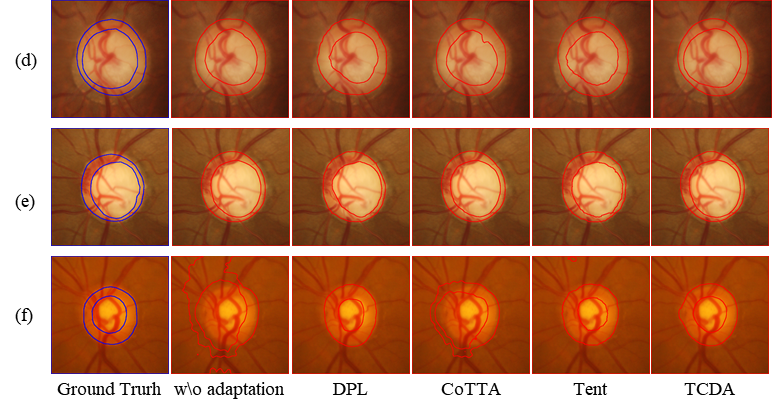}\caption{Visual segmentation results for samples on the Drishti-GS dataset}\label{visD}
	\end{figure}

	To qualitatively evaluate the adaptation performance of different methods, we have visualized the segmentation results on the RIM-ONE-r3 and Drishti-GS datasets, as shown in Fig.\ref{visR} and Fig.\ref{visD}.
	
	By analyzing the results, it can be observed that our method exhibits superior performance for generating more accurate and consistent segmentation results for both the optic disc and optic cup when compared to other methods. It indicates the effectiveness of local boundary consistency constraints in accurately delineating the boundaries of these structures.  Moreover, the combined effect of two consistency constraints ensures the reliability and stability of the segmentation results of our model. In particular, upon observing examples Fig. \ref{visR}(b) and Fig. \ref{visD}(f), we can see even when the source domain model exhibits subpar segmentation performance on the target domain, employing our test-time adaptive training method can notably improve the segmentation performance of the source domain model on target domain images.
	
	\subsection{Ablation Study}
	\subsubsection{Module validity experiments}
	To evaluate the efficacy of our approach, we conduct ablation experiments on the RIM-ONE-r3 dataset and the Drishti-GS dataset. In Table \ref{R消融实验} and Table \ref{D消融实验}, the loss ${L_{Tseg}}$, ${L_{bc}}$, and ${L_{fc}}$ represent the segmentation loss, local boundary consistency loss, and global feature consistency loss, respectively. The symbol "\checkmark" indicates the inclusion of the loss in the experiment, while the symbol "\ding{53}" indicates that the corresponding module is not added to this ablative group. The evaluation metrics used for the ablation experiments are the Dice coefficient and ASD coefficient. Due to the potential catastrophic forgetting and performance degradation associated with directly applying consistency losses for domain adaptive training of the source domain pre-trained model, we first introduce the segmentation loss based on pseudo labels ${L_{Tseg}}$ for domain adaptive training. Subsequently, we sequentially introduce the local boundary consistency loss ${L_{bc}}$ and global feature consistency loss ${L_{fc}}$ in the ablation experiments. The results of these experiments are presented in Table \ref{R消融实验} and Table \ref{D消融实验}.

	\begin{table}[H]
		\renewcommand{\arraystretch}{1.2}
		\centering
		\caption{Ablation experiment results of MCDA model in RIM-ONE-r3 dataset}
		\label{R消融实验}
		\resizebox{\linewidth}{!}{{\begin{tabular}{ccccccccc}
					\hline
					\multirow{2}{*}{${L_{Tseg}}$} & \multirow{2}{*}{${L_{bc}}$}& \multirow{2}{*}{${L_{fc}}$} &\multicolumn{2}{c}{Optic disc segmentation} & \multicolumn{2}{c}{Optic cup segmentation} & \multicolumn{2}{c}{Avg}\\
					& &    & Dice $[\% ]$ & ASD(piexl) & Dice $[\% ]$ & ASD(piexl)& Dice  & ASD\\
					\hline
					\ding{53} & \ding{53} & \ding{53} & 85.96$ \pm $5.32  & 13.48$ \pm $5.57  & 74.95$ \pm $19.04 & 10.58$ \pm $5.87 & 80.46 & 12.02 \\
					\checkmark & \ding{53} & \ding{53} & 85.58$ \pm $7.45  & 15.86$ \pm $10.87 & 78.57$ \pm $18.34 & 9.45$ \pm $7.73  & 82.07 & 12.65 \\
					\checkmark & \checkmark & \ding{53} & 90.32$ \pm $11.53 & 11.52$ \pm $11.64 & 80.61$ \pm $17.85 & 8.26$ \pm $7.46  & 85.46 & 9.89 \\
					\checkmark & \ding{53} & \checkmark & 85.57$ \pm $6.96  & 15.91$ \pm $10.63 & 79.69$ \pm $15.60 & 9.22$ \pm $7.64  & 82.63 & 12.57 \\
					
					\checkmark & \checkmark & \checkmark & \textbf{90.58$ \pm $10.99} & \textbf{10.99$ \pm $9.72}  & \textbf{82.87$ \pm $13.15} & \textbf{7.32$ \pm $5.27}  & \textbf{86.73} & \textbf{9.16} \\ \hline
		\end{tabular}}}
	\end{table}
	\begin{table}[H]
		\renewcommand{\arraystretch}{1.2}
		\centering
		\caption{Ablation experiment results of MCDA model in Drishti-GS dataset}
		\label{D消融实验}
		\resizebox{\linewidth}{!}{{\begin{tabular}{ccccccccc}
					\hline
					\multirow{2}{*}{${L_{Tseg}}$} & \multirow{2}{*}{${L_{bc}}$}& \multirow{2}{*}{${L_{fc}}$} &\multicolumn{2}{c}{Optic disc segmentation} & \multicolumn{2}{c}{Optic cup segmentation} & \multicolumn{2}{c}{Avg}\\
					& &    & Dice $[\% ]$ & ASD(piexl) & Dice $[\% ]$ & ASD(piexl)& Dice  & ASD\\
					\hline
					\ding{53} & \ding{53} & \ding{53} & 95.42$ \pm $1.78 & 5.28$ \pm $2.09 & 85.19$ \pm $12.21 & 9.69$ \pm $5.72 & 90.31 & 7.49 \\
					\checkmark & \ding{53} & \ding{53} & 95.52$ \pm $1.79 & 5.15$ \pm $2.01 & 86.37$ \pm $10.32 & 9.00$ \pm $5.17 & 90.95 & 7.07 \\
					\checkmark & \checkmark & \ding{53} & \textbf{96.14$ \pm $1.76} & \textbf{4.39$ \pm $1.97} & 86.38$ \pm $11.66 & 8.82$ \pm $4.87 & 91.26 & \textbf{6.61} \\
					\checkmark & \ding{53} & \checkmark & 95.57$ \pm $1.69 & 5.09$ \pm $1.98 & \textbf{86.51$ \pm $10.44} & 8.85$ \pm $5.01 & 91.04 & 6.97 \\
					\checkmark & \checkmark & \checkmark & 96.02$ \pm $1.75 & 4.56$ \pm $1.96 & 86.51$ \pm $12.13 & \textbf{8.71$ \pm $5.09} & \textbf{91.27} & 6.64 \\ \hline
		\end{tabular}}}
	\end{table}
	
	Firstly, we introduce the local boundary consistency loss ${L_{bc}}$ on top of the segmentation loss based on pseudo-label ${L_{Tseg}}$. The results are shown in the third row of Table \ref{R消融实验} and Table \ref{D消融实验}. As observed in Table \ref{R消融实验}, on the RIM-ONE-r3 dataset, the Dice coefficients for the optic disc and cup segmentation tasks reach 90.58\% and 82.87\%, respectively. Compared to the results without the introduced ${L_{bc}}$ as shown in the second row of Table \ref{R消融实验}, there is an improvement of 2.04\% and 4.74\%, respectively. Additionally, the ASD coefficients for the optic disc and cup segmentation tasks are lower than those achieved by utilizing only ${L_{Tseg}}$ for domain adaptation training, with reductions of 1.19 and 4.34, respectively. This indicates that introducing the local boundary consistency loss, by aligning the predicted boundaries from the segmentation predictions directly predicting boundaries and exploiting the prior knowledge contained in the source domain model and target domain images for adaptive training, is beneficial in adapting the source pre-trained model to the data distribution of the target domain at the test time. Similar results are also observed in the Drishti-GS dataset, in Table \ref{D消融实验}. Specifically, the Dice score for optic disc segmentation is improved by 0.62\% and the ASD score is decreased by 0.76. Similarly, for optic cup segmentation, the ASD coefficient is decreased by 0.18 when compared to the results obtained without incorporating the target boundary consistency loss as shown in the second row of Table \ref{D消融实验}. These findings highlight the positive impact of this loss for better preserving boundary structures and boosting inter-class discrimination in the Drishti-GS dataset. Overall, the model's performance in optic disc segmentation tasks on both datasets is significantly improved with the incorporation of the local boundary consistency constraint. This improvement can be attributed to the sharp boundaries present in the optic disc. The introduced local boundary consistency constraint proves effective in domain adaptation tasks dealing with domains featuring prominent boundary structures.
	
	The fourth row of Table \ref{R消融实验} and Table \ref{D消融实验} represent the experimental results of introducing the global feature consistency constraint loss ${L_{fc}}$ for domain adaptation based on the segmentation loss ${L_{Tseg}}$. In the Drishti-GS target domain, compared to using only ${L_{Tseg}}$ for adaptation, the Dice coefficients for optic disc and cup segmentation are improved by 0.05\% and 0.14\%, respectively, while the ASD coefficients are reduced by 0.06 and 0.15, respectively. This indicates that the global feature consistency constraint effectively enables the model to focus on the tissue during the adaptive process and enforces the intra-class consistency of global features. Similar performance is observed in the RIM-ONE-r3 dataset, where the introduction of global feature consistency constraint improves cup segmentation accuracy while maintaining optic disc segmentation accuracy. The Dice score for cup segmentation increases from 78.57\% to 79.69\%. Meanwhile, the ASD score for cup segmentation decreases from 9.45 to 9.22. 
	
	Finally, the last row of Table \ref{R消融实验} presents the experimental results of domain adaptation using segmentation loss, local boundary consistency loss, and global feature consistency loss in the RIM-ONE-r3 dataset. For optic disc segmentation, our model achieves a Dice score of 90.58\% and an ASD score of 10.99. Compared to the non-adaptive method as shown in the first row of Table \ref{R消融实验}, the Dice coefficient is improved by 4.62\%, and the ASD coefficient is decreased by 2.49. For cup segmentation, our model achieves a Dice score of 82.87\% and an ASD score of 7.32. Compared to the baseline (non-adaptive), the Dice coefficient is improved by 7.92\%, and the ASD coefficient is decreased by 3.26. Similar results can be found in the Drishti-GS dataset. These results suggest that local boundary consistency and global feature consistency are beneficial for enhancing local inter-class discriminability and global intra-class consistency.
	
	\subsubsection{Ablative analysis of  $\alpha$}
	
\begin{figure}[h]
		\centering\includegraphics[scale=0.25]{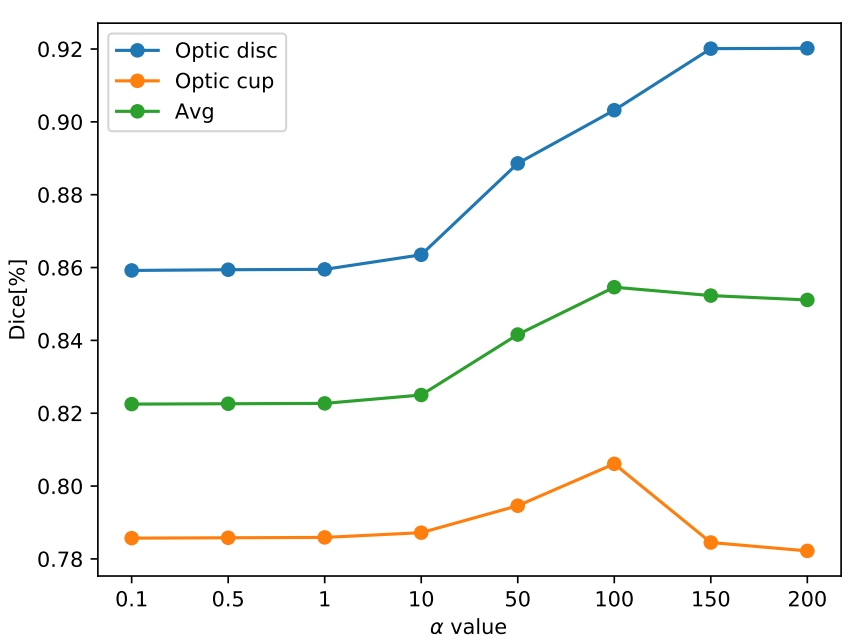}\caption{MCDA model hyperparameter $\alpha $ setting}\label{AA}
	\end{figure}

	In this paper, we propose a local boundary consistency constraint. During the testing phase of the adaptive process, a hyperparameter $\alpha $ is set to control the strength of the consistency task. A larger $\alpha $ value indicates a stronger local boundary consistency constraint, while a smaller value indicates a weaker constraint. To achieve the optimal performance of the model, we conducted ablation experiments on the $\alpha $ hyperparameter on the RIM-ONE-r3 dataset, using the Dice coefficient as the evaluation metric. The experimental results are shown in Table \ref{a消融}. To visually compare the results of different $\alpha $ values, we also plot them in Fig.\ref{AA}.
	\begin{table}[H]
		\renewcommand{\arraystretch}{1}
		\centering
		\caption{Ablative analysis of \textbf{$\alpha$}}
		\label{a消融}
		\resizebox{\linewidth}{!}{{\begin{tabular}{ccccccccc}
					\hline
					$\alpha $   & 0.1   & 0.5   & 1     & 10   & 50    & 100   & 150   & 200   \\
					\hline
					Optic disc[\%] & 85.92 & 85.94 & 85.95 & 86.35 & 88.86 & 90.32 & 92.01 & \textbf{92.02} \\
					Optic cup[\%] & 78.57 & 78.58 & 78.59 & 78.72 & 79.46 & \textbf{80.61} & 78.45 & 78.22 \\
					Avg[\%]  & 82.25 & 82.26 & 82.27 & 82.50 & 84.16 & \textbf{85.46} & 85.23 & 85.11\\
					\hline
		\end{tabular}}}
	\end{table}
	
	Table \ref{a消融} and Fig.\ref{AA} demonstrate that as the value of $\alpha $ increases incrementally, the constraint for consistent boundaries in the tissue edge becomes more stringent, leading to improved performance in segmenting the optic disc. However, if $\alpha $ becomes excessively large, it may cause a decrease in the performance of segmenting the optic cup. A larger value of $\alpha$ indicates a stronger local edge consistency constraint. However, in fundus images, the optic cup has an ambiguous boundary. When dealing with the segmentation of tissue having weak boundaries, excessive reliance on the target edge consistency constraint leads to unreliable segmentation results. We set the value of $\alpha $ at 100 according to the ablative study.
	
	\subsubsection{Ablative analysis of $\beta $}
	
\begin{figure}[h]
		\centering\includegraphics[scale=0.25]{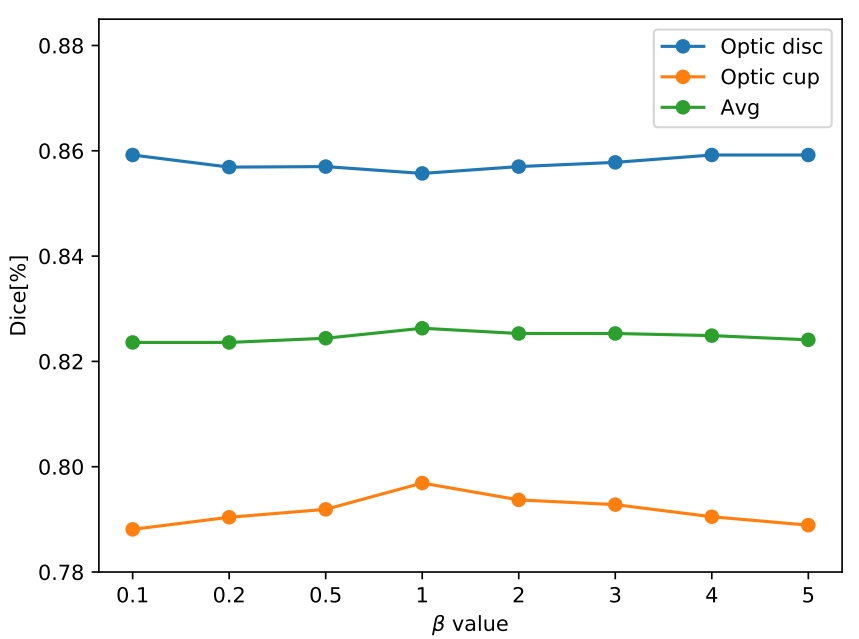}\caption{MCDA model hyperparameter $\beta $ setting}\label{BB}
	\end{figure}

	During the test time of the adaptive process, the hyperparameter $\beta $ is set to control the strength of the consistency task. A larger $\beta $ value indicates a stronger global feature consistency constraint, while a smaller value indicates a weaker constraint. In order to achieve the optimal performance of the model, we conduct ablation experiments on the $\beta $ hyperparameter on the RIM-ONE-r3 dataset, using the Dice coefficient as the evaluation metric. The experimental results are shown in Table \ref{b消融}. To visually compare the results of different $\beta $ values, we plot them in Fig.\ref{BB}.

	\begin{table}[H]
		\renewcommand{\arraystretch}{1}
		\centering
		\caption{Ablative analysis of \textbf{$\beta $}}
		\label{b消融}
		\resizebox{\linewidth}{!}{{\begin{tabular}{ccccccccc}
					\hline
					$\beta $  & 0.1    & 0.2    & 0.5    & 1      & 2      & 3     & 4     & 5     \\
					\hline
					Optic disc[\%] & 85.92  & 85.69  & 85.70   & 85.57  & 85.70   & 85.78 & \textbf{85.92} & 85.92 \\
					Optic cup[\%] & 78.81  & 79.04  & 79.19  & \textbf{79.69}  & 79.37  & 79.28 & 79.05 & 78.89 \\
					Avg[\%]  & 82.36 & 82.36 & 82.44 & \textbf{82.63} & 82.53 & 82.53 & 82.49 & 82.41\\
					\hline
		\end{tabular}}}
	\end{table}
	
	As shown in Table \ref{b消融} and Fig.\ref{BB}, it can be observed that the Dice coefficient for optic disc segmentation fluctuates within a certain range for different values of $\beta $, while the Dice coefficient for optic cup segmentation reaches its highest value when $\beta $ is set to 1. In order to balance the segmentation results of the optic cup and disc, we set the value of $\beta $ as 1.

	\section{Conclusions}
	\label{section7}
	In this paper, we propose a multi task consistency guided test-time source-free medical image segmentation method. We introduce both a local boundary consistency constraint and a global feature consistency constraint, aiming to offer a suitable consistency signal for test-time domain adaptation. These constraints are proved to benefit to boosting local inter-class discriminability and global inter-class consistency. 
	We conduct extensive experiments on the fundus image segmentation task. The experimental results demonstrate that the proposed MCDA exhibits superior performance compared to other competitive algorithms across all metrics. In future work, toward personalized medicine, we will explore adaptive methods for testing only a single image, allowing the model to provide only a single image segmentation results for each test image.

	\bibliography{mybibfile}

\begin{thebibliography}{10}
\expandafter\ifx\csname url\endcsname\relax
  \def\url#1{\texttt{#1}}\fi
\expandafter\ifx\csname urlprefix\endcsname\relax\def\urlprefix{URL }\fi
\expandafter\ifx\csname href\endcsname\relax
  \def\href#1#2{#2} \def\path#1{#1}\fi

\bibitem{sun2020test}
Y.~Sun, X.~Wang, Z.~Liu, J.~Miller, A.~Efros, M.~Hardt, Test-time training with
  self-supervision for generalization under distribution shifts, in:
  International Conference on Machine Learning, 2020, pp. 9229--9248.

\bibitem{chen2022contrastive}
D.~Chen, D.~Wang, T.~Darrell, S.~Ebrahimi, Contrastive test-time adaptation,
  in: Proceedings of the IEEE/CVF Conference on Computer Vision and Pattern
  Recognition, 2022, pp. 295--305.

\bibitem{ma2022test}
W.~Ma, C.~Chen, S.~Zheng, J.~Qin, H.~Zhang, Q.~Dou, Test-time adaptation with
  calibration of medical image classification nets for label distribution
  shift, in: Medical Image Computing and Computer Assisted Intervention, 2022,
  pp. 313--323.

\bibitem{bateson2022test}
M.~Bateson, H.~Lombaert, I.~Ben~Ayed, Test-time adaptation with shape moments
  for image segmentation, in: Medical Image Computing and Computer Assisted
  Intervention, 2022, pp. 736--745.

\bibitem{karani2021test}
N.~Karani, E.~Erdil, K.~Chaitanya, E.~Konukoglu, Test-time adaptable neural
  networks for robust medical image segmentation, Medical Image Analysis (2021)
  101907.

\bibitem{yang2022dltta}
H.~Yang, C.~Chen, M.~Jiang, Q.~Liu, J.~Cao, P.~A. Heng, Q.~Dou, Dltta: Dynamic
  learning rate for test-time adaptation on cross-domain medical images, IEEE
  Transactions on Medical Imaging (2022) 3575--3586.

\bibitem{wang2022continual}
Q.~Wang, O.~Fink, L.~Van~Gool, D.~Dai, Continual test-time domain adaptation,
  in: Proceedings of the IEEE/CVF Conference on Computer Vision and Pattern
  Recognition, 2022, pp. 7201--7211.

\bibitem{yu2019uncertainty}
L.~Yu, S.~Wang, X.~Li, C.-W. Fu, P.-A. Heng, Uncertainty-aware self-ensembling
  model for semi-supervised 3d left atrium segmentation, in: Medical Image
  Computing and Computer Assisted Intervention, 2019, pp. 605--613.

\bibitem{ouali2020semi}
Y.~Ouali, C.~Hudelot, M.~Tami, Semi-supervised semantic segmentation with
  cross-consistency training, in: Proceedings of the IEEE/CVF Conference on
  Computer Vision and Pattern Recognition, 2020, pp. 12674--12684.

\bibitem{li2020transformation}
X.~Li, L.~Yu, H.~Chen, C.-W. Fu, L.~Xing, P.-A. Heng, Transformation-consistent
  self-ensembling model for semisupervised medical image segmentation, IEEE
  Transactions on Neural Networks and Learning Systems (2020) 523--534.

\bibitem{kurmi2021domain}
V.~K. Kurmi, V.~K. Subramanian, V.~P. Namboodiri, Domain impression: A source
  data free domain adaptation method, in: Proceedings of the IEEE/CVF winter
  Conference on Applications of Computer Vision, 2021, pp. 615--625.

\bibitem{kim2021domain}
Y.~Kim, D.~Cho, K.~Han, P.~Panda, S.~Hong, Domain adaptation without source
  data, IEEE Transactions on Artificial Intelligence (2021) 508--518.

\bibitem{yang2022source}
C.~Yang, X.~Guo, Z.~Chen, Y.~Yuan, Source free domain adaptation for medical
  image segmentation with fourier style mining, Medical Image Analysis (2022)
  102457.

\bibitem{ye2021source}
M.~Ye, J.~Zhang, J.~Ouyang, D.~Yuan, Source data-free unsupervised domain
  adaptation for semantic segmentation, in: Proceedings of the 29th ACM
  International Conference on Multimedia, 2021, pp. 2233--2242.

\bibitem{chen2021source}
C.~Chen, Q.~Liu, Y.~Jin, Q.~Dou, P.-A. Heng, Source-free domain adaptive fundus
  image segmentation with denoised pseudo-labeling, in: Medical Image Computing
  and Computer Assisted Intervention, 2021, pp. 225--235.

\bibitem{xu2022denoising}
Z.~Xu, D.~Lu, Y.~Wang, J.~Luo, D.~Wei, Y.~Zheng, R.~K.-y. Tong, Denoising for
  relaxing: Unsupervised domain adaptive fundus image segmentation without
  source data, in: Medical Image Computing and Computer Assisted Intervention,
  2022, pp. 214--224.

\bibitem{vs2022target}
V.~VS, J.~M.~J. Valanarasu, V.~M.Patel, Target and task specific source-free
  domain adaptive image segmentation, arXiv preprint arXiv:2203.15792 (2022).

\bibitem{fleuret2021uncertainty}
F.~Fleuret, et~al., Uncertainty reduction for model adaptation in semantic
  segmentation, in: Proceedings of the IEEE/CVF Conference on Computer Vision
  and Pattern Recognition, 2021, pp. 9613--9623.

\bibitem{bateson2020source}
M.~Bateson, H.~Kervadec, J.~Dolz, H.~Lombaert, I.~Ben~Ayed, Source-relaxed
  domain adaptation for image segmentation, in: Medical Image Computing and
  Computer Assisted Intervention, 2020, pp. 490--499.

\bibitem{he2020momentum}
K.~He, H.~Fan, Y.~Wu, S.~Xie, R.~Girshick, Momentum contrast for unsupervised
  visual representation learning, in: Proceedings of the IEEE/CVF Conference on
  Computer Vision and Pattern Recognition, 2020, pp. 9729--9738.

\bibitem{liu2022vmfnet}
X.~Liu, S.~Thermos, P.~Sanchez, A.~Q. O’Neil, S.~A. Tsaftaris, vmfnet:
  Compositionality meets domain-generalised segmentation, in: Medical Image
  Computing and Computer Assisted Intervention, 2022, pp. 704--714.

\bibitem{wang2020tent}
D.~Wang, E.~Shelhamer, S.~Liu, B.~Olshausen, T.~Darrell, Tent: Fully test-time
  adaptation by entropy minimization, arXiv preprint arXiv:2006.10726 (2020).

\bibitem{guo2022joint}
X.~Guo, J.~Li, Q.~Lin, Z.~Tu, X.~Hu, S.~Che, Joint optic disc and cup
  segmentation using feature fusion and attention, Computers in Biology and
  Medicine (2022) 106094.

\bibitem{li2021medical}
S.~Li, X.~Sui, X.~Luo, X.~Xu, Y.~Liu, R.~Goh, Medical image segmentation using
  squeeze-and-expansion transformers, arXiv preprint arXiv:2105.09511 (2021).

\bibitem{kervadec2019boundary}
H.~Kervadec, J.~Bouchtiba, C.~Desrosiers, E.~Granger, J.~Dolz, I.~B. Ayed,
  Boundary loss for highly unbalanced segmentation, in: International
  Conference on Medical Imaging with Deep Learning, 2019, pp. 285--296.

\bibitem{chen2021deep}
X.~Chen, X.~Luo, G.~Wangy, Y.~Zhengy, Deep elastica for image segmentation, in:
  2021 IEEE 18th International Symposium on Biomedical Imaging (ISBI), 2021,
  pp. 706--710.

\bibitem{orlando2020refuge}
J.~I. Orlando, H.~Fu, J.~B. Breda, K.~Van~Keer, D.~R. Bathula, A.~Diaz-Pinto,
  R.~Fang, P.-A. Heng, J.~Kim, J.~Lee, et~al., Refuge challenge: A unified
  framework for evaluating automated methods for glaucoma assessment from
  fundus photographs, Medical Image Analysis (2020) 101570.

\bibitem{fumero2011rim}
F.~Fumero, S.~Alay{\'o}n, J.~L. Sanchez, J.~Sigut, M.~Gonzalez-Hernandez,
  Rim-one: An open retinal image database for optic nerve evaluation, in: 2011
  24th International Symposium on Computer-Based Medical Systems, 2011, pp.
  1--6.

\bibitem{sivaswamy2015comprehensive}
J.~Sivaswamy, S.~Krishnadas, A.~Chakravarty, G.~Joshi, A.~S. Tabish, et~al., A
  comprehensive retinal image dataset for the assessment of glaucoma from the
  optic nerve head analysis, JSM Biomedical Imaging Data Papers (2015) 1004.

\bibitem{wang2019boundary}
S.~Wang, L.~Yu, K.~Li, X.~Yang, C.-W. Fu, P.-A. Heng, Boundary and
  entropy-driven adversarial learning for fundus image segmentation, in:
  Medical Image Computing and Computer Assisted Intervention, 2019, pp.
  102--110.

\bibitem{vu2019advent}
T.-H. Vu, H.~Jain, M.~Bucher, M.~Cord, P.~P{\'e}rez, Advent: Adversarial
  entropy minimization for domain adaptation in semantic segmentation, in:
  Proceedings of the IEEE/CVF Conference on Computer Vision and Pattern
  Recognition, 2019, pp. 2517--2526.

\end{thebibliography}

\end{document}